\documentclass[onecolumn]{IEEEtran}
\usepackage{spconf,amsmath,graphicx,mathtools,amssymb,units, color}
\usepackage[numbers,sort&compress]{natbib}
\usepackage{flushend}
\usepackage{algorithm}
\usepackage[noend]{algpseudocode}

\title{Enhancing ICA performance by exploiting sparsity: Application to \\ fMRI Analysis}
\name{Zois~Boukouvalas$^1$, Yuri Levin-Schwartz$^2$, and T\"{u}lay Adal{\i}$^2$}
\address{ $^1$University of Maryland, Baltimore County, Dept. of Mathematics and Statistics, Baltimore, MD 21250\\
$^2$University of Maryland, Baltimore County, Dept. of CSEE, Baltimore, MD 21250 }

\begin{document}
%
\maketitle
\begin{abstract}
Independent component analysis (ICA) is a powerful method for blind source separation based on the assumption that sources are statistically independent. Though ICA has proven useful and has been employed in many applications, complete statistical independence can be too restrictive an assumption in practice. Additionally, important prior information about the data, such as sparsity, is usually available. Sparsity is a natural property of the data, a form of diversity, which, if  incorporated into the ICA model, can relax the independence assumption, resulting in an improvement in the overall separation performance. In this work, we propose a new variant of ICA by entropy bound minimization (ICA-EBM)---a flexible, yet parameter-free algorithm---through the direct exploitation of sparsity. Using this new SparseICA-EBM algorithm, we study the synergy of independence and sparsity through simulations on synthetic as well as functional magnetic resonance imaging (fMRI)-like data.

\end{abstract}
\section{Introduction}
\label{sec:intro}
Independent component analysis (ICA) is a data-driven method that provides a unique decomposition of a dataset solely through the assumption that sources are statistically independent and has found wide use in a variety of applications. Although statistical independence is a natural assumption in many cases, there are many practical applications where such a strong assumption is unrealistic. Often in these cases, some important prior information about the data is available and incorporating it into the ICA model will result in better overall separation performance.


A widely used approach for incorporating prior information into the ICA framework is through the use of constrained independent component analysis (C-ICA) \cite{lu2005approach}, which incorporates prior information using equality and inequality constraints under a Lagrangian framework. Such prior information can be about the task in functional magnetic resonance imaging fMRI analysis and can be included as constraints on the mixing matrix columns \cite{calhoun2005semi,6960099,wang2011fixed} or spatial maps  \cite{lin2010semiblind,soldati2013ica,lu2005approach}. While this approach is practical, such constraints have to be in an exact functional form, something that is not always available in practice. Another form of prior information that can be considered are natural properties of the data, such as sparsity. 


There are many ways to impose sparsity into the ICA model, such as by selecting a density model that favors sparse distributions \cite{bell1995information,hyvarinen1999fast} or by using sparsity transformations \cite{5495320} following ICA. Although, selecting the source distribution would allow the ICA model to enjoy the desirable large sample properties of the ML formulation \cite{6784026, Comon_HandbookBSS_2010}, the model would be limited to a specific type of sparse distribution \cite{6784026}. Additionally, sparsity transformations are an indirect way of imposing sparsity and do not allow direct way to control independence versus sparsity.



The main contribution of this work is the development of a new ICA algorithm that takes the sparsity of each individual source into account.  We incorporate sparsity into the ICA algorithm, entropy bound minimization (ICA-EBM) \cite{li2010independent}, by introducing a weighting factor to the ICA cost function to balance the contribution of sparsity for each of the individual sources. ICA-EBM is a flexible and parameter-free ICA algorithm that can separate sources from a wide range of distributions. The new SparseICA-EBM algorithm inherits all the advantages of ICA-EBM, namely its flexibility, though with enhanced performance due to the exploitation of the sparsity and allows the user to balance the roles of independence and sparsity.



The remainder of this paper is organized as follows. In Section 2, we provide the necessary background on ICA. Section 3, provides the mathematical development of SparseICA-EBM. In Section 4, we demonstrate the effectiveness of SpaseICA-EBM through sparse simulated data as well as simulated fMRI-like data. The conclusions are presented in Section 5.

\section{Background}
\label{sec:backgr}
\subsection{Independent Component Analysis}

Let $N$ statistically independent sources ${\bf s}(t) = [ s_1(t),\dots,s_N(t)]^{\top}$ be mixed through an unknown invertible mixing matrix ${\bf A}\in \mathbb{R}^{N\times N}$ so that we obtain mixtures ${\bf x}(t) = [ x_1(t),\dots,x_N(t)]^{\top}$, through the linear model 
\begin{equation}\nonumber
{\bf x}(t) = {\bf A}{\bf s}(t),~~~ t=1,\dots,T,
\end{equation}
where $t$ denotes the discrete time index and $(\cdot)^{\top}$ the transpose. The goal of ICA is to estimate a demixing matrix ${\bf W}\in \mathbb{R}^{N\times N}$ to yield maximally independent source estimates ${\bf y}(t) = {\bf W}{\bf x}(t)$. A natural cost function to achieve such a separation is mutual information (MI), which is defined as the Kullback-Leibler (KL)-distance between the joint source density and the product of the marginal estimated source densities. Therefore, the MI cost function is given by
\begin{equation}\label{ICAmutualInfo}
J_{ICA}({\bf W}) = \sum_{n=1}^N H(y_n) - \log|\det({\bf W})| - H({\bf x}),
\end{equation}
where $y_n = {\bf w}_n^{\top} {\bf x}$ and the terms $H(y_n)$ and $H({\bf x})$ are the (differential) entropy of the source estimates and the mixtures, respectively. Note that the term $H({\bf x})$ is independent of ${\bf W}$ and can be treated as a constant $C$. The minimization of the MI is equivalent to the maximization of the maximum likelihood (ML) cost function, hence, making available all the theoretical advantages associated with the ML theory \cite{6784026} for large sample sizes.

It is impractical to try to exploit prior information in (\ref{ICAmutualInfo}) as it requires either complete knowledge of the demixing matrix or of the sources, information that is not usually available. Assuming that the demixing matrix is orthogonal would loosen this strict requirement, but would unecessarily limit the solution space. Moreover, direct implementation of (\ref{ICAmutualInfo}) implies that each latent source has the same distribution, which is unrealistic in many practical applications. All of these issues can be avoided by rewriting (\ref{ICAmutualInfo}) and its gradient with respect to each row of ${\bf W}$, ${\bf w}_m,\ m=1,\dots N$. Thus, by using this decoupling approach \cite{li2007nonorthogonal,li2010independent}, the MI cost function can be written as 
\begin{equation}\label{ICAmutualInfoDecoupling}
J_{ICA}({\bf w}_m) = \sum_{n=1}^NH(y_n) - \log\left| {\bf h}_m^{\top} {\bf w}_m \right| - C_m,~m = 1,\dotsc,N, 
\end{equation}
where ${\bf h}_m$ is a unit vector that is perpendicular to all row vectors of ${\bf W}$ except ${\bf w}_m$ and $C_m$ is a constant. The gradient of (\ref{ICAmutualInfoDecoupling}) can be written in a decoupled form and is given by 
\begin{equation}\label{gradcost}
\frac{\partial J_{ICA}({\bf w}_m)}{\partial {\bf w}_m} = -E\left\{ \phi(y_m) {\bf x} \right\} - \frac{{\bf h}_m}{ {\bf h}_m^{\top} {\bf w}_m},\nonumber
\end{equation} 
where $\phi(y_m) = \frac{{\partial \log p(y_m)}}{{\partial y_m}}$ is called the score function and the probability density function (PDF) of the $m$th estimated source, $p(y_m)$, can be adaptively determined for each estimated source independently.

\section{Mathematical Development and Implementation}

The formal definition for sparsity is given through the $\ell^0$ norm and is defined as the number of non-zero coefficients from a vector $y\in \mathbb{R}^T$
\begin{equation}\label{SparsityL0Norm}
||y||_0 = \# \{y_i\neq 0;j=1,\dots T\}.
\end{equation}
Although the incorporation of (\ref{SparsityL0Norm}) into the ICA framework is a direct way to impose sparsity, the $\ell^0$ norm is computationally intractable. Instead, the $\ell^1$ norm, defined as the sum of the absolute values of a vector's coefficients, serves as a surrogate sparsity regularizer of the $\ell^0$ norm in many optimization problems \cite{candes2005decoding,schmidt2007fast,tibshirani1996regression}.
We can promote the synergy between independence and sparsity through the addition of the $\ell^1$ regularization term to (\ref{ICAmutualInfoDecoupling}), which we expect to improve separation performance when the underlying sources are truly sparse.

The proposed decoupled sparsity promoting ICA cost function is thus given by
\begin{equation}\label{SparseCostFunction}
J({\bf w}_m) = J_{ICA}({\bf w}_m) + \lambda_m f(y_m),~m = 1,\dotsc,N,
\end{equation}
where $f(y_m) = ||y_m||_1$ is the regularization term and $\lambda_m$ is called the sparsity parameter. The $\ell^1$ norm is a non-differentiable function, so it is replaced by the the sum of multi-quadratic functions \cite{lee2006efficient}, given by 
\begin{equation}
f(y_m) =  \lim_{\epsilon\to 0} \sum_{t=1}^T\sqrt{y_{m_t}^2 + \epsilon},\nonumber
\end{equation}
where $\epsilon$ is the smoothing parameter. Therefore, the proposed gradient can be written as
\begin{align*}
		\frac{\partial}{\partial {\bf w}_m} J({\bf w}_m)&=\frac{\partial}{\partial {\bf w}_m} \left(J_{ICA}({\bf w}_m) + \lambda_m \lim_{\epsilon\to 0} f(y_m)\right)\\ \\ 
		&= \frac{\partial J_{ICA}({\bf w}_m)}{\partial {\bf w}_m} + \lambda_m \lim_{\epsilon\to 0}\sum_{t=1}^T \frac{y_{m_t}}{\sqrt{y_{m_t}^2 + \epsilon}}{\bf x}.
\end{align*}

Due to its ability to maximize independence in an efficient manner through the use of four measuring functions favoring bimodal, symmetric or skewed, heavy-tailed or not heavy-tailed distributions \cite{li2010independent}, ICA-EBM serves as the algorithm for the direct integration of (\ref{SparseCostFunction}). The new SparseICA-EBM not only provides flexible density matching but also yields solutions with variable levels of sparsity.

\section{Experimental Results}
\label{sec:mathdev}

We demonstrate the performance of SparseICA-EBM (\ref{SparseCostFunction}), in terms of its separation power, using simulated sparse sources as well as simulated fMRI-like data. We compare the SparseICA-EBM algorithm with the original ICA-EBM algorithm. Additionally, due to its popularity in many applications including fMRI analysis, we also compare SparseICA-EBM with two implementations of the Infomax algorithm \cite{bell1995information}. One version is based on the natural gradient optimization framework (Infomax-NG) and the other one is based on a quasi-Newton technique Broyden, Fletcher, Goldfarb, and Shanno (BFGS) \cite{nielsen2000ucminf}, which we call Infomax-BFGS. The hardware used in the computational studies is part of the UMBC High Performance Computing Facility (HPCF), for more information see hpcf.umbc.edu.

\subsection{Simulated Sparse Sources}
For the first set of experiments, we generate 20 simulated sources, each of which is distributed according to a generalized Gaussian distribution (GGD) with sample size $T=10^3$. The PDF of each source is given by \cite{nadarajah2005generalized}
\begin{equation}
p(x;\beta, \sigma) = \eta\exp\left( -\frac{x}{2\sigma} \right)^{2\beta},\ x\in\mathbb{R} \nonumber
\end{equation}
where $\eta = \frac{\beta}{2^{\frac{1}{2\beta}}\Gamma(\frac{1}{2\beta})\sigma}$. The shape parameter, $\beta$, controls the peakedness and spread of the distribution as well as its sparsity. If $\beta<1$, the distribution is more peaky than the Gaussian with heavier tails, and if $\beta > 1$, it is less peaky with lighter tails. Thus, as $\beta\to 0$ the distribution becomes more sparse. 

\begin{figure}[H]	
	\begin{minipage}[b]{1.0\linewidth}
		\centering
		\centerline{\includegraphics[width=7cm]{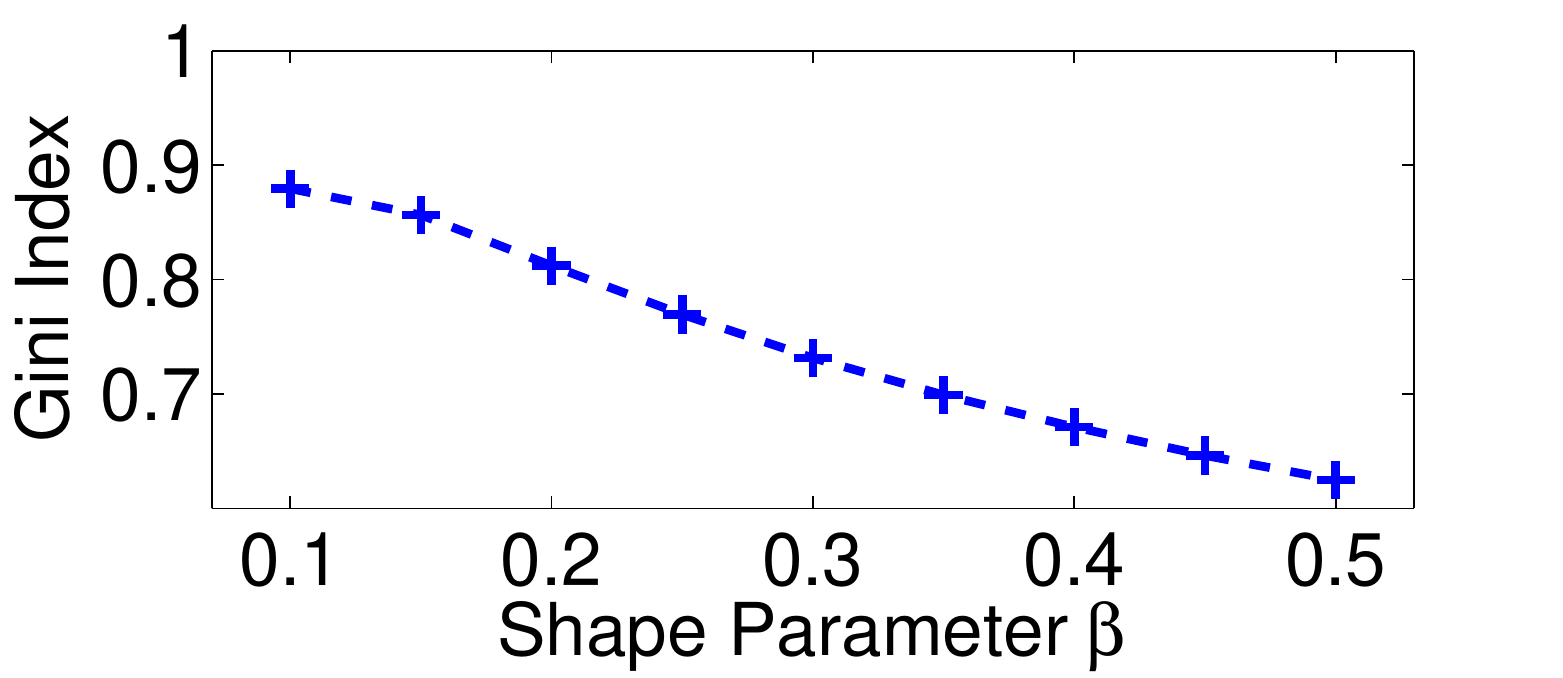}}
		\vspace{-8mm}
	\end{minipage}
	\hfill
	\caption{Average Gini Index as a function of the shape parameter, $\beta$. The Gini Index is normalized and 1 corresponds to very sparse sources while 0 to dense sources.}
	\label{GiniIndex}
\end{figure}

To verify the sparse nature of the sources used for the first set of the experiments, we generate 20 sources with sample size $T=10^4$ and shape parameter $\beta$ from the range $[0.1,0.5]$ with a step size of $0.05$. For each specific source, we measure the sparsity level using the Gini Index as described in \cite{hurley2009comparing} and average over the sources that correspond to a specific $\beta$. In Fig.~\ref{GiniIndex}, we see that as we increase $\beta$, sources become less sparse.
\vspace{-3.5mm}

To evaluate the performance of the algorithms, we use the average-interference-to-signal ratio (ISR) as in \cite{li2007nonorthogonal}. For SparseICA-EBM, the algorithm parameters are $\lambda = 10^4$ and $\epsilon = 10^{-2}$ and are determined based on a grid search selection. All results are the average of 300 independent runs.

\begin{figure}[htb]	
	\begin{minipage}[b]{1.0\linewidth}
		\centering
		\centerline{\includegraphics[width=7.5cm]{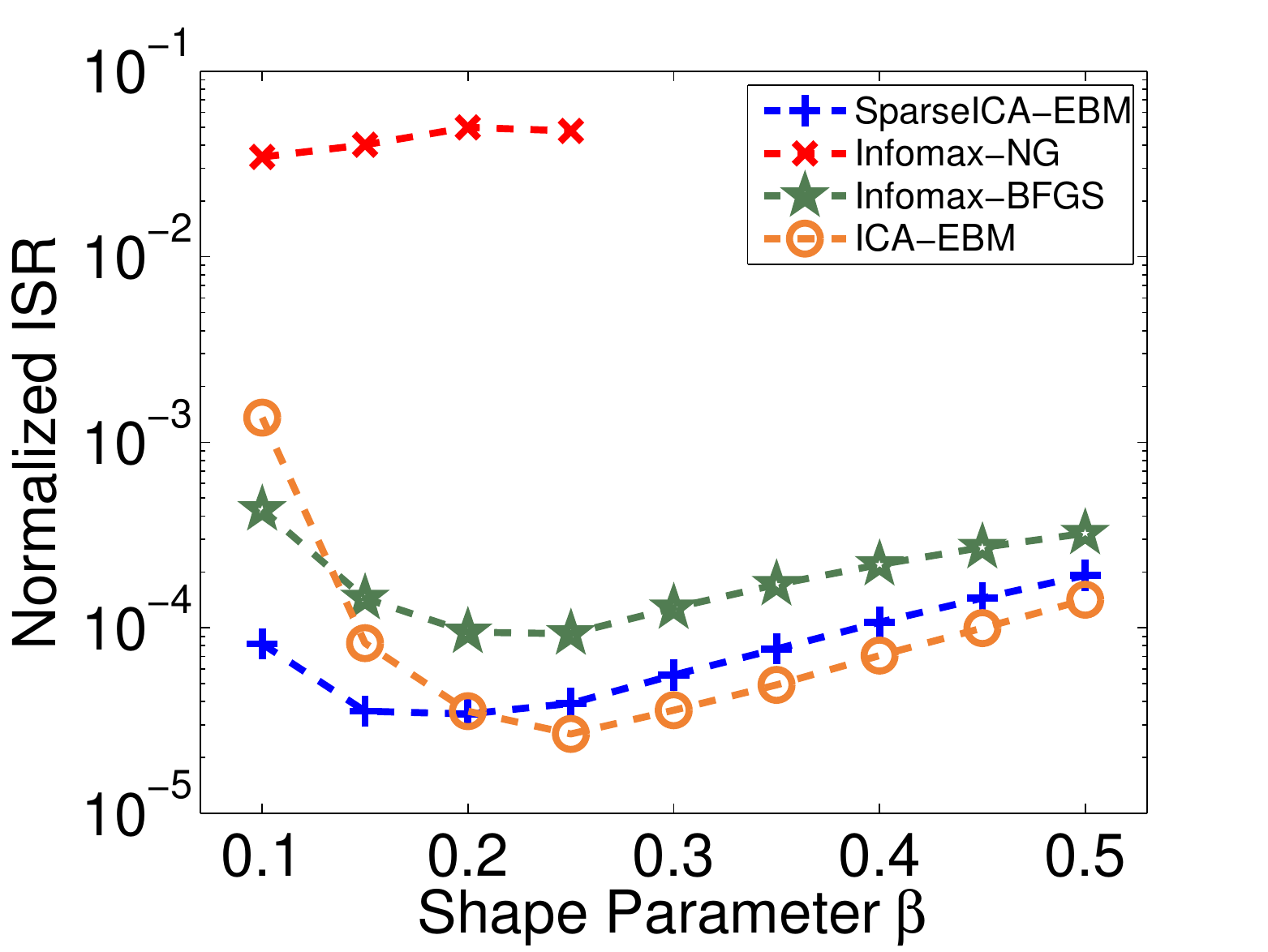}}
		\vspace{-8mm}
	\end{minipage}
	\hfill
	\caption{Performance comparison of four ICA algorithms in terms of the normalized average ISR as a function of shape parameter, $\beta$, for 20 sources with $T=10^3$. Each point is the result of 300 independent runs.}
	\label{wrtbeta}
\end{figure}

In Fig.~\ref{wrtbeta}, we display the normalized ISR as a function of $\beta$. We observe that for small values of $\beta$, {\it i.e}., highly sparse case, SparseICA-EBM exhibits better performance. On the other hand, ICA-EBM starts performing better than the other algorithms as we increase $\beta$, {\it i.e.}, decrease sparsity. It is worth mentioning that Infomax-NG often fails to converge as $\beta$ increases revealing its poor performance under this experimental setup. On the other hand, Infomax-BFGS shows reasonable performance especially for small values of $\beta$.
\vspace{-2mm}

\begin{figure}[htb]	
	\begin{minipage}[b]{1.0\linewidth}
		\centering
		\centerline{\includegraphics[width=7.5cm]{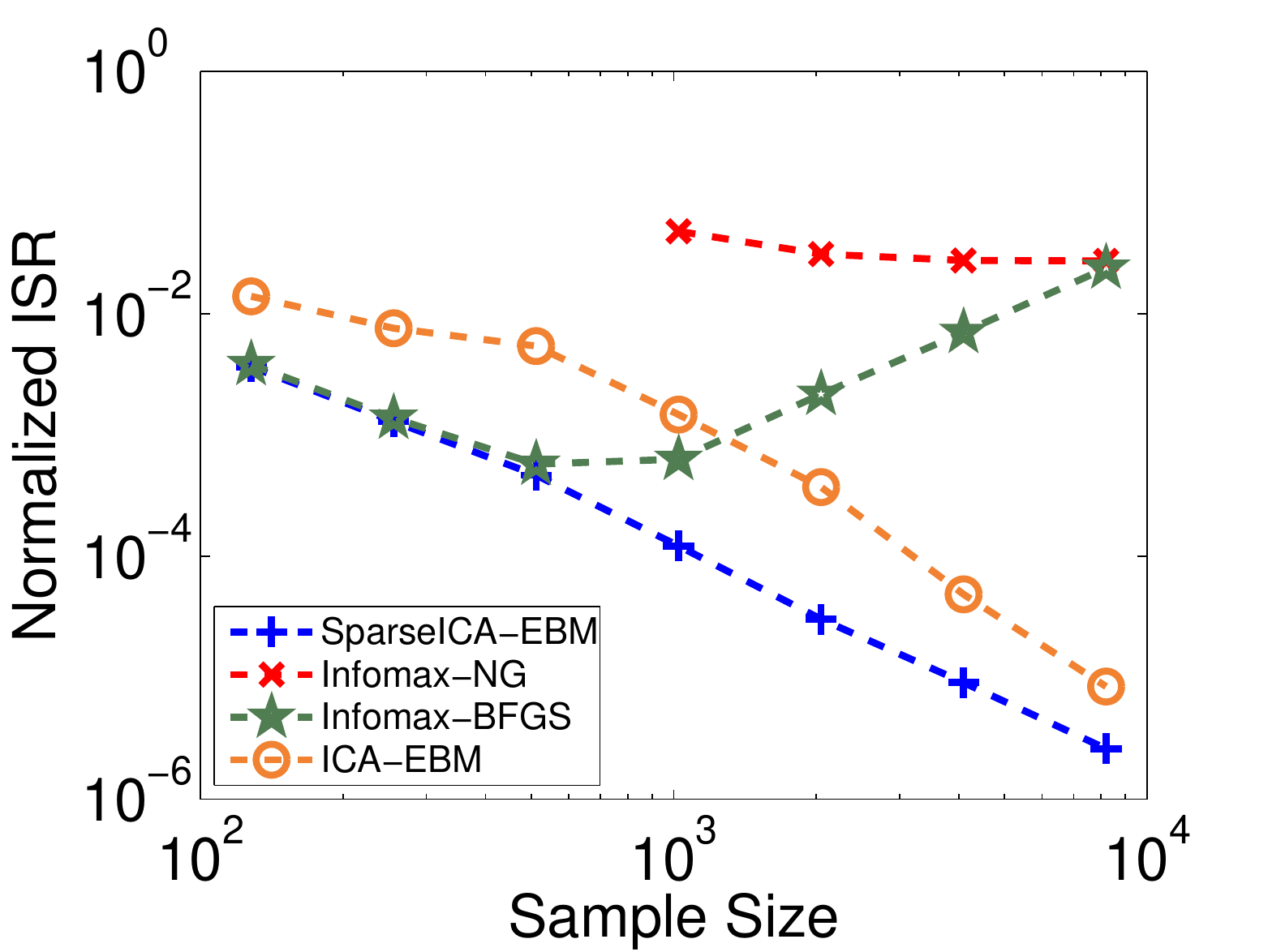}}
		\vspace{-9mm}
	\end{minipage}
	\hfill
	\caption{Performance comparison of four ICA algorithms in terms of the normalized average ISR as a function of sample size, $T$, for 20 sources with $\beta=0.1$. Each point is the result of 300 independent runs.}
	\label{wrtSamples}
\end{figure}

In Fig.~\ref{wrtSamples}, we display the normalized ISR as a function of the sample size. To study the case where sources are very sparse we generate all sources using $\beta=0.1$. As the sample size increases, SparseICA-EBM and ICA-EBM perform better than the other two algorithms, since the large sample size enables an accurate approximation of the differential entropy of the estimated sources. When the sample size becomes greater than $10^3$, Infomax-BFGS starts providing highly inaccurate results, due to algorithmic issues in the approximation of the inverse of the Hessian matrix.
\vspace{-2mm}

\begin{figure}[htb]	
	\begin{minipage}[b]{1.0\linewidth}
		\centering
		\centerline{\includegraphics[width=7.5cm]{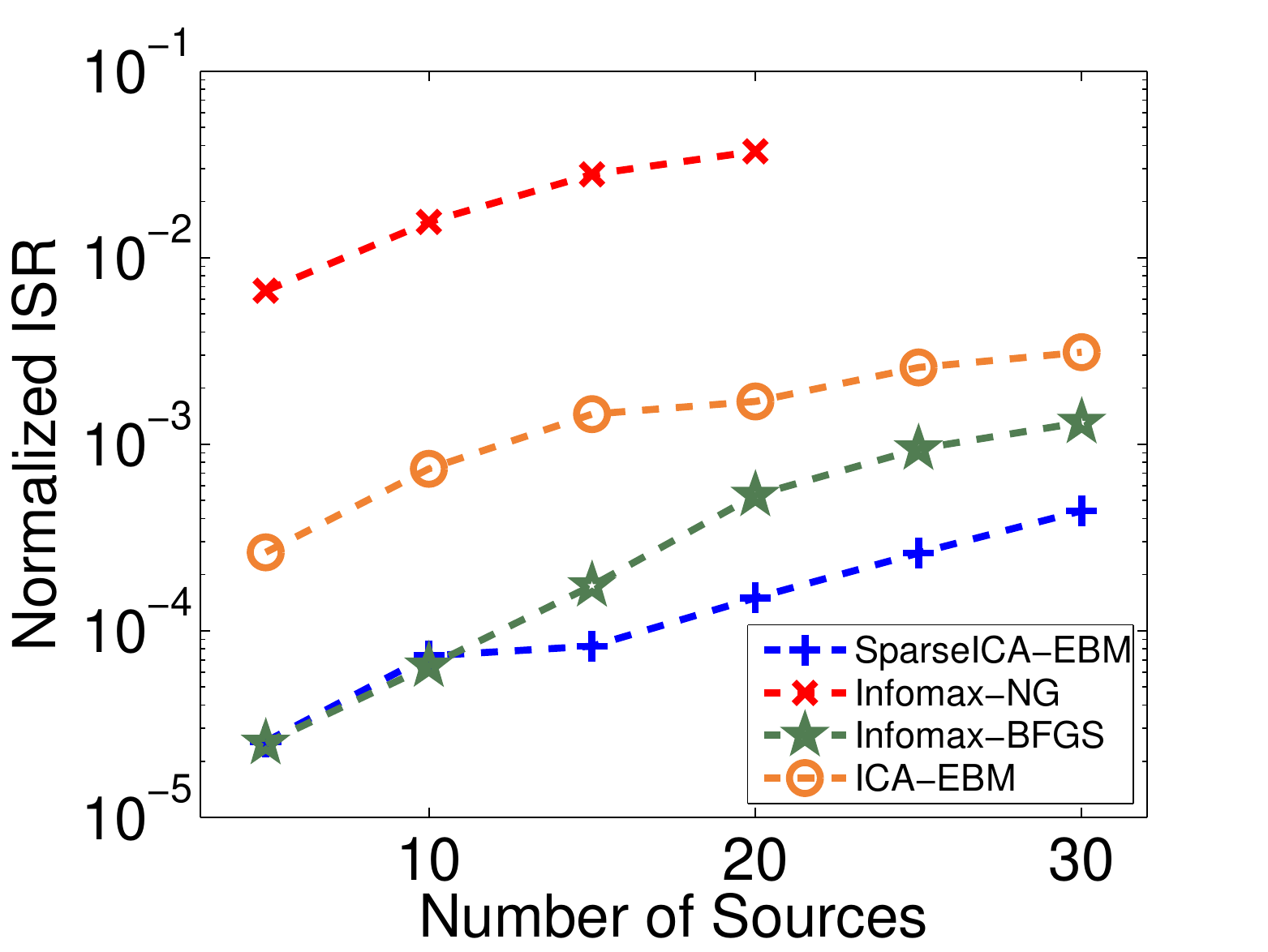}}
		\vspace{-9mm}
	\end{minipage}
	\hfill
	\caption{Performance comparison of four ICA algorithms in terms of the normalized average ISR as a function of number of sources, $N$, with $T=10^3$ and $\beta=0.1$. Each point is the result of 300 independent runs.}
	\label{wrtSources}
\end{figure}

Finally, in Fig.~\ref{wrtSources}, we display the normalized ISR as a function of the number of sources where for each source $T=10^3$ and $\beta=0.1$. It is clear from Fig. \ref{wrtSources} that SparseICA-EBM shows the best performance. Infomax-BFGS performs well when the number of sources is small since the optimization procedure is performed in a low dimensional space. This reveals the benefit of employing the decoupling approach, since the reduction to a set of vector optimization problems avoids over-complicated surfaces for the cost function. 
\vspace{-2mm}

\vspace{-4mm}
\subsection{Simulated fMRI-like Data}
For the second set of experiments, we used simulated fMRI-like sources at different contrast-to-noise ratio (CNR) levels. Estimated spatial fMRI sources tend to have sparse distributions \cite{6268324}, leading to the popularity of sparsity
favoring algorithms such as Infomax. Since Infomax-NG performed poorly for all the experiments using simulated sources, for this set of experiments, we only compare SparseICA-EBM with the original ICA-EBM and Infomax-BFGS. To generate fMRI-like data we use the fMRI toolbox, SimTB \cite{erhardt2012simtb}, which enables flexible generation of fMRI datasets under a model of spatiotemporal separability.

For our experiment, we generate 20 spatial maps using 10 subjects. Each spatial mask is a $100\times100$ image with a baseline intensity of 800. The length of the experiment is 260 samples. Rician noise is added to each dataset at specified CNR value. The parameters for SparseICA-EBM, $\lambda = 0.31$ and $\epsilon = 10^{-1}$, are determined based on a grid search selection method performed on noiseless data.

\begin{figure}[htb]	
	\begin{minipage}[b]{1.0\linewidth}
		\centering
		\centerline{\includegraphics[width=4cm]{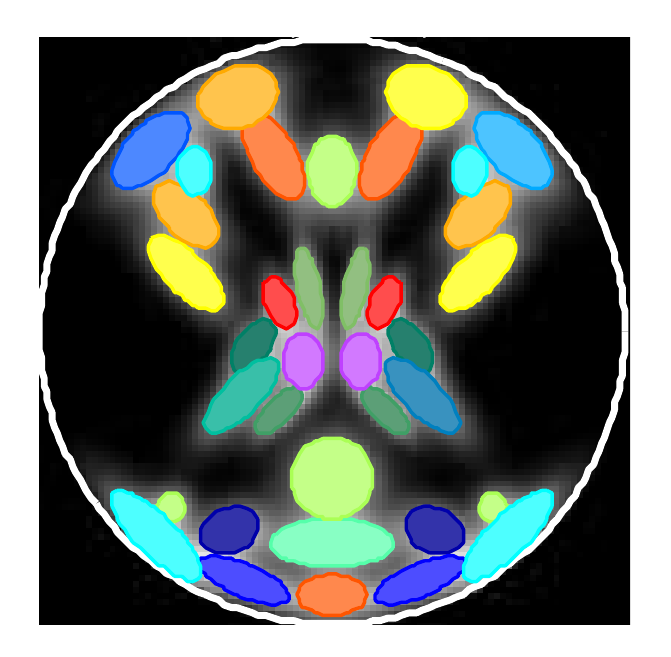}}
		\vspace{-9mm}
	\end{minipage}
	\hfill
	\caption{Simulated fMRI-like components. Note that each color indicates a different component.}
	\label{Components}
\end{figure}

The first step in processing the fMRI like-data consists of the application of principal component analysis to each dataset individually. Since 20 sources are generated for each dataset, the dimension of each dataset is reduced to 20. After dimension reduction, we apply the ICA algorithms to each dataset, such that we are seeking spatially independent components that correspond to spatial functional connectivity maps shown in Fig.~\ref{Components}. After obtaining the estimated demixing matrices from each of the algorithms and for each dataset, we estimate the independent components and, together with their associated demixing vectors, pair them with the true sources.  In the case where more than one estimated component is paired with a single true source, we use Bertsekas algorithm \cite{bertsekas1988auction} to find the best assignment. To evaluate the performance of the ICA algorithms, we use the average absolute value of the correlation between the true and the estimated sources.  
\vspace{-2mm}

\begin{figure}[htb]	
	\begin{minipage}[b]{1.0\linewidth}
		\centering
		\centerline{\includegraphics[width=7.5cm]{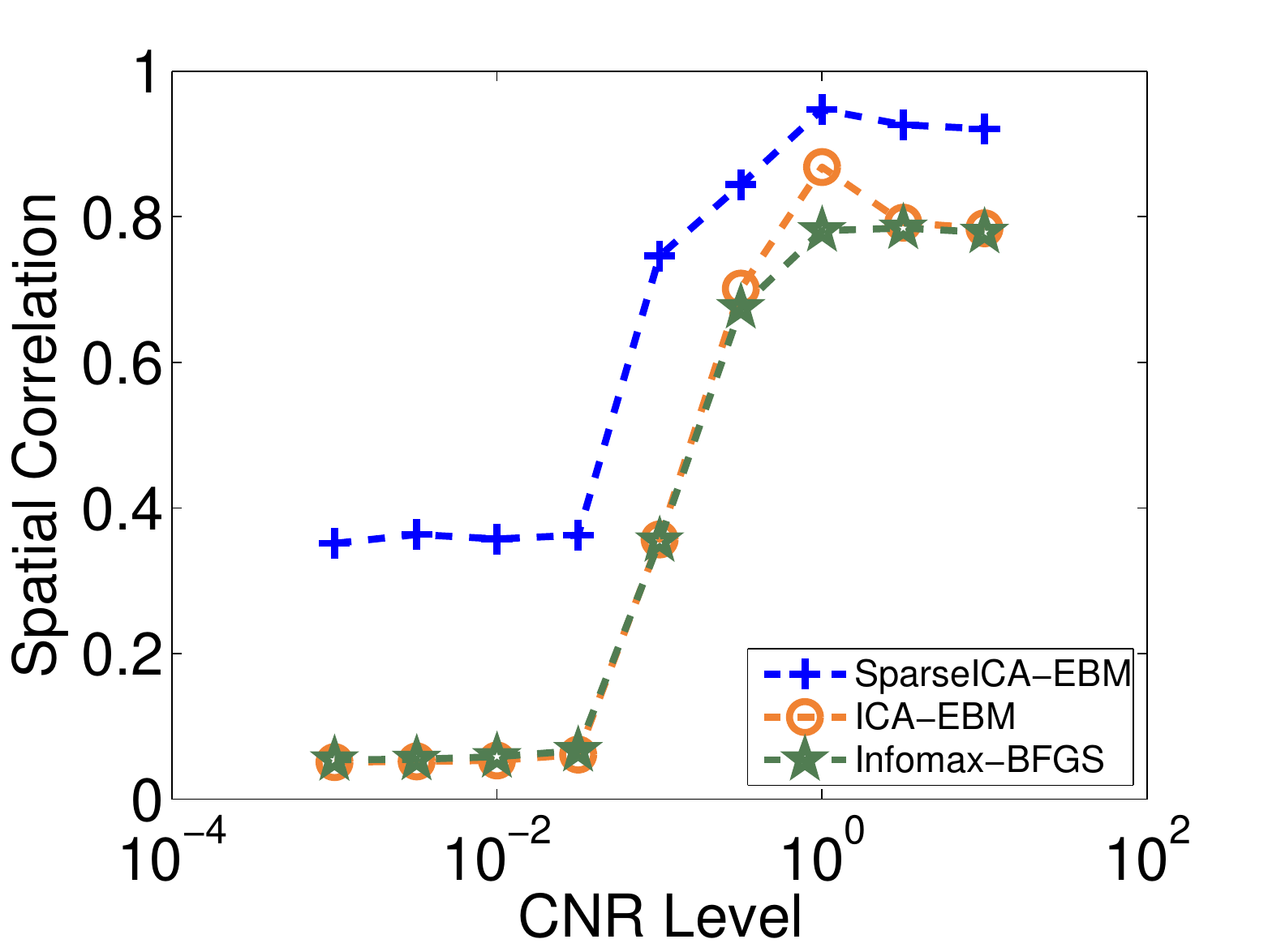}}
		\vspace{-9mm}
	\end{minipage}
	\hfill
	\caption{Spatial correlation between the true and the estimated sources as a function of the CNR level. Each point is the result of 300 independent runs.}
	\label{fMRI}
\end{figure}

As expected, we observe in Fig.~\ref{fMRI}, that SparseICA-EBM provides superior performance compared with the other algorithms. Specifically, we note a nearly $100\%$ improvement in performance for CNR values below 0.1. We conclude that SparseICA-EBM is more robust to noise than classical ICA algorithms as it incorporates true prior information of the sources into the estimation.
\vspace{-2mm}

\vspace{-5mm}
\section{Conclusion}
\label{sec:conc}
Both sparsity and independence have proven useful in a variety of applications, motivating the development of an algorithm that can effectively balance the contributions of these two forms of diversity. In this paper, we propose a new ICA algorithm, SparseICA-EBM, that takes the sparsity of each individual source directly into account. Experimental results, using synthetic as well as simulated fMRI-like data, confirm the superiority of SparseICA-EBM over traditional ICA algorithms. This, combined with the fact that SparseICA-EBM is robust to noise, make it an attractive ICA algorithm for applications where prior information about the sparsity of the sources is available.


\vfill\pagebreak

\bibliographystyle{IEEEtran}
\bibliography{./ReferencesICASSP17}

\begin{thebibliography}{10}
\providecommand{\url}[1]{#1}
\csname url@samestyle\endcsname
\providecommand{\newblock}{\relax}
\providecommand{\bibinfo}[2]{#2}
\providecommand{\BIBentrySTDinterwordspacing}{\spaceskip=0pt\relax}
\providecommand{\BIBentryALTinterwordstretchfactor}{4}
\providecommand{\BIBentryALTinterwordspacing}{\spaceskip=\fontdimen2\font plus
\BIBentryALTinterwordstretchfactor\fontdimen3\font minus
  \fontdimen4\font\relax}
\providecommand{\BIBforeignlanguage}[2]{{%
\expandafter\ifx\csname l@#1\endcsname\relax
\typeout{** WARNING: IEEEtran.bst: No hyphenation pattern has been}%
\typeout{** loaded for the language `#1'. Using the pattern for}%
\typeout{** the default language instead.}%
\else
\language=\csname l@#1\endcsname
\fi
#2}}
\providecommand{\BIBdecl}{\relax}
\BIBdecl

\bibitem{lu2005approach}
W.~Lu and J.~C. Rajapakse, ``Approach and applications of constrained {ICA},''
  \emph{IEEE transactions on neural networks}, vol.~16, no.~1, pp. 203--212,
  2005.

\bibitem{calhoun2005semi}
V.~Calhoun, T.~Adal{\i}, M.~Stevens, K.~Kiehl, and J.~Pekar, ``Semi-blind {ICA}
  of f{MRI}: a method for utilizing hypothesis-derived time courses in a
  spatial {ICA} analysis,'' \emph{Neuroimage}, vol.~25, no.~2, pp. 527--538,
  2005.

\bibitem{6960099}
P.~A. Rodriguez, M.~Anderson, V.~D. Calhoun, and T.~Adal{\i}, ``General
  nonunitary constrained {ICA} and its application to complex-valued f{MRI}
  data,'' \emph{IEEE Transactions on Biomedical Engineering}, vol.~62, no.~3,
  pp. 922--929, March 2015.

\bibitem{wang2011fixed}
Z.~Wang, ``Fixed-point algorithms for constrained {ICA} and their applications
  in f{MRI} data analysis,'' \emph{Magnetic resonance imaging}, vol.~29, no.~9,
  pp. 1288--1303, 2011.

\bibitem{lin2010semiblind}
Q.-H. Lin, J.~Liu, Y.-R. Zheng, H.~Liang, and V.~D. Calhoun, ``Semiblind
  spatial {ICA} of f{MRI} using spatial constraints,'' \emph{Human brain
  mapping}, vol.~31, no.~7, pp. 1076--1088, 2010.

\bibitem{soldati2013ica}
N.~Soldati, V.~D. Calhoun, L.~Bruzzone, and J.~Jovicich, ``{ICA} analysis of
  f{MRI} with real-time constraints: an evaluation of fast detection
  performance as function of algorithms, parameters and a priori conditions,''
  \emph{Frontiers in human neuroscience}, vol.~7, p.~19, 2013.

\bibitem{bell1995information}
A.~J. Bell and T.~J. Sejnowski, ``An information-maximization approach to blind
  separation and blind deconvolution,'' \emph{Neural computation}, vol.~7,
  no.~6, pp. 1129--1159, 1995.

\bibitem{hyvarinen1999fast}
A.~Hyvarinen, ``Fast and robust fixed-point algorithms for independent
  component analysis,'' \emph{IEEE transactions on Neural Networks}, vol.~10,
  no.~3, pp. 626--634, 1999.

\bibitem{5495320}
S.~Ma, X.~L. Li, N.~M. Correa, T.~Adal{\i}, and V.~D. Calhoun, ``Independent
  subspace analysis with prior information for f{MRI} data,'' in \emph{2010
  IEEE International Conference on Acoustics, Speech and Signal Processing},
  March 2010, pp. 1922--1925.

\bibitem{6784026}
T.~Adal{\i}, M.~Anderson, and G.-S. Fu, ``Diversity in independent component
  and vector analyses: Identifiability, algorithms, and applications in medical
  imaging,'' \emph{IEEE Signal Processing Magazine}, vol.~31, no.~3, pp.
  18--33, May 2014.

\bibitem{Comon_HandbookBSS_2010}
P.~Comon and C.~Jutten, \emph{Handbook of Blind Source Separation: Independent
  Component Analysis and Applications}, 1st~ed.\hskip 1em plus 0.5em minus
  0.4em\relax Academic Press, 2010.

\bibitem{li2010independent}
X.-L. Li and T.~Adal{\i}, ``Independent component analysis by entropy bound
  minimization,'' \emph{IEEE Trans. Signal Processing}, vol.~58, no.~10, pp.
  5151--5164, 2010.

\bibitem{li2007nonorthogonal}
X.-L. Li and X.-D. Zhang, ``Nonorthogonal joint diagonalization free of
  degenerate solution,'' \emph{IEEE Trans. Signal Processing}, vol.~55, no.~5,
  pp. 1803--1814, 2007.

\bibitem{candes2005decoding}
E.~J. Candes and T.~Tao, ``Decoding by linear programming,'' \emph{IEEE
  transactions on information theory}, vol.~51, no.~12, pp. 4203--4215, 2005.

\bibitem{schmidt2007fast}
M.~Schmidt, G.~Fung, and R.~Rosales, ``Fast optimization methods for $\ell^1$
  regularization: A comparative study and two new approaches,'' in
  \emph{European Conference on Machine Learning}.\hskip 1em plus 0.5em minus
  0.4em\relax Springer, 2007, pp. 286--297.

\bibitem{tibshirani1996regression}
R.~Tibshirani, ``Regression shrinkage and selection via the {LASSO},''
  \emph{Journal of the Royal Statistical Society. Series B (Methodological)},
  pp. 267--288, 1996.

\bibitem{lee2006efficient}
S.-I. Lee, H.~Lee, P.~Abbeel, and A.~Y. Ng, ``Efficient $\ell^1$ 1 regularized
  logistic regression,'' in \emph{Proceedings of the National Conference on
  Artificial Intelligence}, vol.~21, no.~1.\hskip 1em plus 0.5em minus
  0.4em\relax Menlo Park, CA; Cambridge, MA; London; AAAI Press; MIT Press;
  1999, 2006, p. 401.

\bibitem{nielsen2000ucminf}
H.~B. Nielsen, ``Ucminf-an algorithm for unconstrained, nonlinear
  optimization,'' Informatics and Mathematical Modelling (IMM), Technical
  University of Denmark, Tech. Rep., 2000.

\bibitem{nadarajah2005generalized}
S.~Nadarajah, ``A generalized normal distribution,'' \emph{Journal of Applied
  Statistics}, vol.~32, no.~7, pp. 685--694, 2005.

\bibitem{hurley2009comparing}
N.~Hurley and S.~Rickard, ``Comparing measures of sparsity,'' \emph{IEEE
  Transactions on Information Theory}, vol.~55, no.~10, pp. 4723--4741, 2009.

\bibitem{6268324}
V.~D. Calhoun and T.~Adal{\i}, ``Multisubject independent component analysis of
  f{MRI}: A decade of intrinsic networks, default mode, and neurodiagnostic
  discovery,'' \emph{IEEE Reviews in Biomedical Engineering}, vol.~5, pp.
  60--73, 2012.

\bibitem{erhardt2012simtb}
E.~B. Erhardt, E.~A. Allen, Y.~Wei, T.~Eichele, and V.~D. Calhoun, ``Sim{TB}, a
  simulation toolbox for f{MRI} data under a model of spatiotemporal
  separability,'' \emph{Neuroimage}, vol.~59, no.~4, pp. 4160--4167, 2012.

\bibitem{bertsekas1988auction}
D.~P. Bertsekas, ``The auction algorithm: A distributed relaxation method for
  the assignment problem,'' \emph{Annals of operations research}, vol.~14,
  no.~1, pp. 105--123, 1988.

\end{thebibliography}

\end{document}